\documentclass[twocolumn,10pt,a4paper]{article}

\title{Deep Bayesian Active Semi-Supervised Learning}
\author{Matthias Rottmann\begin{NoHyper}\thanks{Bergische Universiat\"at Wuppertal, Faculty of Mathematics and Natural Sciences, \texttt{\{rottmann,kkahl\}@math.uni-wuppertal.de}, \texttt{hanno.gottschalk@uni-wuppertal.de}}\end{NoHyper}, \ Karsten Kahl\footnotemark[1] \ and Hanno Gottschalk\footnotemark[1]}

\date{}

\usepackage[utf8]{inputenc}
\usepackage[english]{babel}
\usepackage{amsmath,amsfonts,amsthm}

\usepackage{algorithm}
\usepackage[vlined,boxruled,linesnumbered,algo2e,algosection]{algorithm2e}
\usepackage{graphicx}
\usepackage{subfigure}
\usepackage{placeins}
\usepackage{float}
\usepackage{xcolor}
\usepackage{tikz}
\usepackage[colorlinks=true,
citecolor=black!50!green,
linkcolor=black!33!red,
]{hyperref}
\usepackage[
style=authoryear,
giveninits,
url=false,
doi=false,
isbn=false,
eprint=false,
mincitenames=1,
maxcitenames=1,
uniquelist=false]{biblatex}

\DeclareCiteCommand{\cite}
  {\usebibmacro{prenote}}
  {\usebibmacro{citeindex}%
   \printtext[bibhyperref]{\usebibmacro{cite}}}
  {\multicitedelim}
  {\usebibmacro{postnote}}

\DeclareCiteCommand*{\cite}
  {\usebibmacro{prenote}}
  {\usebibmacro{citeindex}%
   \printtext[bibhyperref]{\usebibmacro{citeyear}}}
  {\multicitedelim}
  {\usebibmacro{postnote}}

\DeclareCiteCommand{\parencite}[\mkbibparens]
  {\usebibmacro{prenote}}
  {\usebibmacro{citeindex}%
    \printtext[bibhyperref]{\usebibmacro{cite}}}
  {\multicitedelim}
  {\usebibmacro{postnote}}

\DeclareCiteCommand*{\parencite}[\mkbibparens]
  {\usebibmacro{prenote}}
  {\usebibmacro{citeindex}%
    \printtext[bibhyperref]{\usebibmacro{citeyear}}}
  {\multicitedelim}
  {\usebibmacro{postnote}}

\DeclareCiteCommand{\footcite}[\mkbibfootnote]
  {\usebibmacro{prenote}}
  {\usebibmacro{citeindex}%
  \printtext[bibhyperref]{ \usebibmacro{cite}}}
  {\multicitedelim}
  {\usebibmacro{postnote}}

\DeclareCiteCommand{\footcitetext}[\mkbibfootnotetext]
  {\usebibmacro{prenote}}
  {\usebibmacro{citeindex}%
   \printtext[bibhyperref]{\usebibmacro{cite}}}
  {\multicitedelim}
  {\usebibmacro{postnote}}

\DeclareCiteCommand{\textcite}
  {\boolfalse{cbx:parens}}
  {\usebibmacro{citeindex}%
   \printtext[bibhyperref]{\usebibmacro{textcite}}}
  {\ifbool{cbx:parens}
     {\bibcloseparen\global\boolfalse{cbx:parens}}
     {}%
   \multicitedelim}
  {\usebibmacro{textcite:postnote}}

\usepackage{csquotes}
\usepackage[nameinlink]{cleveref}
\usetikzlibrary{intersections,plotmarks}

\usepackage{titlesec}
\titleformat*{\section}{\large\bfseries}
\titlespacing*{\section}{0pt}{0.8\baselineskip}{0.2\baselineskip}
\titlespacing*{\subsection}{0pt}{0.8\baselineskip}{0.2\baselineskip}
\titlespacing*{\paragraph}{0pt}{0.4\baselineskip}{0.4\baselineskip}

\renewbibmacro{in:}{%
  \ifboolexpr{%
     test {\ifentrytype{article}}%
     or
     test {\ifentrytype{inproceedings}}%
  }{}{\printtext{\bibstring{in}\intitlepunct}}%
}

\DeclareFieldFormat{titlecase}{\MakeTitleCase{#1}}

\newrobustcmd{\MakeTitleCase}[1]{%
  \ifthenelse{\ifcurrentfield{booktitle}\OR\ifcurrentfield{booksubtitle}%
    \OR\ifcurrentfield{maintitle}\OR\ifcurrentfield{mainsubtitle}%
    \OR\ifcurrentfield{journaltitle}\OR\ifcurrentfield{journalsubtitle}%
    \OR\ifcurrentfield{issuetitle}\OR\ifcurrentfield{issuesubtitle}%
    \OR\ifentrytype{book}\OR\ifentrytype{mvbook}\OR\ifentrytype{bookinbook}%
    \OR\ifentrytype{booklet}\OR\ifentrytype{suppbook}%
    \OR\ifentrytype{collection}\OR\ifentrytype{mvcollection}%
    \OR\ifentrytype{suppcollection}\OR\ifentrytype{manual}%
    \OR\ifentrytype{periodical}\OR\ifentrytype{suppperiodical}%
    \OR\ifentrytype{proceedings}\OR\ifentrytype{mvproceedings}%
    \OR\ifentrytype{reference}\OR\ifentrytype{mvreference}%
    \OR\ifentrytype{report}\OR\ifentrytype{thesis}}
    {#1}
    {\MakeSentenceCase{#1}}
}

\DeclareFieldFormat[article]{titlecase}{\MakeSentenceCase{#1}}

\theoremstyle{plain}

\theoremstyle{definition}

\theoremstyle{remark}

\newcommand{\SetAlgorithmStyle}{
  \SetKwData{Left}{left}\SetKwData{This}{this}\SetKwData{Up}{up}
  \SetKwInOut{Input}{Input}\SetKwInOut{Output}{Output}
  \ResetInOut{input}
  \SetKwComment{tcp}{//}{}
  \SetKwFor{For}{for}{}{end}
  \SetArgSty{}
  \DontPrintSemicolon
}

\crefalias{AlgoLine}{line}
\crefalias{algocf}{algorithm}

\makeatletter
\let\cref@old@stepcounter\stepcounter
\def\stepcounter#1{%
  \cref@old@stepcounter{#1}%
  \cref@constructprefix{#1}{\cref@result}%
  \@ifundefined{cref@#1@alias}%
    {\def\@tempa{#1}}%
    {\def\@tempa{\csname cref@#1@alias\endcsname}}%
  \protected@edef\cref@currentlabel{%
    [\@tempa][\arabic{#1}][\cref@result]%
    \csname p@#1\endcsname\csname the#1\endcsname}}
\makeatother

\makeatletter
\renewcommand{\algocf@caption@boxruled}{%
  \hrule
  \hbox to \hsize{%
    \vrule\hskip-0.4pt
    \vbox{   
       \vskip\interspacetitleboxruled%
       \unhbox\algocf@capbox\hfill
       \vskip\interspacetitleboxruled
       }%
     \hskip-0.4pt\vrule%
   }\nointerlineskip%
}%
\makeatother

\usepackage{enumerate}
\usepackage{titling}
\begin{document}

\maketitle

\begin{abstract}
    In many applications the process of generating label information is expensive and time consuming. We present a new method that combines active and semi-supervised deep learning to achieve high generalization performance from a deep convolutional neural network with as few known labels as possible. In a setting where a small amount of labeled data as well as a large amount of unlabeled data is available, our method first learns the labeled data set. This initialization is followed by an expectation maximization algorithm, where further training reduces classification entropy on the unlabeled data by targeting a low entropy fit which is consistent with the labeled data. In addition the algorithm asks at a specified frequency an oracle for labels of data with entropy above a certain entropy quantile. Using this active learning component we obtain an agile labeling process that achieves high accuracy, but requires only a small amount of known labels. For the MNIST dataset we report an error rate of $2.06\%$ using only $300$ labels and $1.06\%$ for $1,\!000$ labels. These results are obtained without employing any special network architecture or data augmentation.
\end{abstract}

\section{Introduction}

In recent years deep learning has shown great potential in solving classification and regression tasks of increasing complexity and difficulty. For academic purposes, several labeled data sets with associated tasks are available to support and facilitate research on machine learning.
Though in many practical applications (e.g.~in industry, medicine and microbiology) where raw data is available in abundance, labeled information is not readily available and the process of generating labels can be time consuming and expensive. Therefore, the development of methods that provide strong predictive models from as few labels as possible is a field of high interest.

The fields of active learning and semi-supervised learning address this issue and provide two approaches to obtain strong predictive models using only few labels, see~\cite{Gal:2016:DBA:3045390.3045502,DBLP:conf/icml/HuMTMS17,DBLP:journals/corr/KingmaB14,Lee_pseudo-label:the,10.1007/978-3-662-44851-9_36,DBLP:RasmusVHBR15,NIPS2011_4409,Weston2012}. They both assume a situation where the complete set of data is large, but labels are known only for a small fraction of it. 

The field of semi-supervised learning has a long history. Already in \cite{10.1007/3-540-52255-7_33} unlabeled data had been injected into the training of neural networks in order to improve generalization performance. Most approaches rely on the Expectation Maximization (cf.~\cite{Dempster77maximumlikelihood}, EM) technique which is a clustering algorithm. In the semi-supervised context, EM is used to assign unlabeled data to a finite number of clusters which are initially defined by the small set of labeled samples. That is, an initial model is trained and then, using the resulting model, 
labels are assigned to unlabeled data, which in turn are used to further train the model. 
The pseudo-label approach, introduced in~\cite{Lee_pseudo-label:the}, is such an EM technique. It uses the labels predicted by the neural network itself and can be viewed as well as an auxiliary loss in the training phase which is inserted to reduce classification entropy on the unlabeled data. It is well known that the EM strategy works well in presence of low density class separation. Thus it is unclear if and how this approach is able to adequately classify samples with high classification uncertainty. In case a quantitative measure of classification uncertainty can be defined, unlabeled data should only be used for training  if their uncertainty is small. 
However, some samples typically retain high uncertainty in the semi-supervised training cycle. 
This is where active learning comes into play, in that it is most valuable to acquire ground truth labels for samples with high classification uncertainty and add those to training. In this way active and semi-supervised learning complement each other naturally.

Both learning approaches benefit from good uncertainty quantification mechanisms. With the advent of Monte-Carlo (MC) dropout~\cite{Gal:2016:DBA:3045390.3045502}, we have an instrument at hand that makes it feasible to construct sensitive metrics to monitor classification uncertainty. Bayesian inference has been used in an active deep learning approach introduced in~\cite{DBLP:journals/corr/GalIG17}.

In recent years, there were also efforts on designing specialized network architectures that incorporate components like denoising auto-encoders, see \cite{DBLP:RasmusVHBR15}. Also deep generative models were used for semi-supervised learning, see \cite{DBLP:journals/corr/KingmaRMW14}.

Combining the active learning and the semi-supervised learning track, a method for synthetic aperture radar image recognition has been published in~\cite{DBLP:journals/cin/GaoYWSYZ17}.

In this paper, we present a deep Bayesian Active Semi-Supervised learning (deepBASS) approach that is based on an EM deep learning approach for classification tasks paired with an active learning component and approximate Bayesian uncertainty. We first train a Convolutional Neural Network (CNN) on a small sample of labeled training data. Afterwards we employ the EM technique, i.e., we iteratively predict classes and assign these as pseudo-labels to the unlabeled data set. Then, we train one epoch on the pseudo-labeled data and the ground-truth-labeled data. While doing so, we make sure that the prediction accuracy on the ground truth remains high. During this process, the algorithm asks an oracle for additional label information where the neural network shows increased classification uncertainty, e.g., high classification entropy. For all predictions and uncertainty estimations we incorporate MC dropout inference.

The remainder of this work is structured as follows:
In \cref{sec:relwo} we classify our method with respect to existing approaches in the literature. 
Then we introduce our method in detail in \cref{sec:assdl} including all necessary notations. Using a simple toy example in \cref{sec:exp_illu} we motivate the combination of active and semi-supervised learning. Using the MNIST dataset, we compare two settings in \cref{sec:exp} where on one hand all unlabeled data is present in training from the beginning and where on the other hand unlabeled data is added only incrementally. Both settings are combined with two different label acquisition policies. Concluding the experiments we compare our method with other semi-supervised and active learning approaches. 

\section{Related Work} \label{sec:relwo}

Our aim is to provide concepts of how to train models with high predictive power with as small labeling effort as possible. In this we combine components from active learning, semi-supervised learning and approximate Bayesian uncertainty quantification and construct a robust method that achieves high accuracy. 

A related semi-supervised deep learning method including MC dropout inference has been published in~\cite{DBLP:journals/corr/abs-1710-00209} which incrementally assigns labels to data with highest predicted class probability above a chosen threshold and adds the respective data to the training data, but does not facilitate an active learning components.

On the other hand, an active deep learning approach making use of Bayesian uncertainty has been introduced in~\cite{DBLP:journals/corr/GalIG17}. This work stresses the importance of approximate Bayesian model uncertainty in active learning and shows comparisons with semi-supervised methods. However, this approach does not make use of semi-supervised learning.

In~\cite{DBLP:RasmusVHBR15}, a so-called ladder network with denoising components has been introduced, which achieved outstanding results for the MNIST dataset. This specialized network architecture, however, is not trivially generalizable to more complex tasks, like e.g.\ object classification/detection and semantic segmentation, where state-of-the-art networks are huge. Deep generative models have been employed successfully in semi-supervised learning, but suffer from scalability issues as well, see \cite{DBLP:journals/corr/KingmaRMW14}. The aim of the presented approach is to show that a combination of active and semi-supervised learning techniques is able to achieve similar performance with a much simpler and scalable network architecture.

The active semi-supervised learning approach introduced in \cite{DBLP:journals/cin/GaoYWSYZ17}, a method for synthetic aperture radar image recognition, accepts in every iteration a chosen number of pseudo-labels with highest confidence and asks an oracle for a chosen number of samples with lowest confidence. Confidence is measured in terms of highest classification probability, approximate Bayesian uncertainty is not employed in this approach. We observe in our tests that compared to the average classification entropy of the available initial ground-truth-labeled data, plenty of unlabeled data have classification entropy below this threshold. Therefore, in the beginning, many thousands of unlabeled samples can be automatically labeled and added to training while producing only a tiny fraction of false positives, i.e., incorrect labels. Furthermore, we also address the question whether it is necessary to pseudo-label and add unlabeled data incrementally or at once.



\section{DeepBASS Learning} \label{sec:assdl}
In order to introduce the deep Bayesian active semi-supervised learning approach we first review its basic components.
 
\subsection{Expectation Maximization}

The Expectation Maximization (\cite{Dempster77maximumlikelihood}, EM) algorithm is a widely used clustering approach. In the original unsupervised context this clustering algorithm is initialized on a model with a predefined number of classes and random parameters. When doing semi-supervised learning, this random initial model is replaced by a model that is trained on the scarce ground truth labels. The second phase is always unsupervised and applies the model to all unlabeled (later pseudo-labeled) data in order to cluster it. Here, the clustering metric is provided by the neural network itself via classification entropy. This in turn can be viewed as adding an additional term to the loss function. Afterwards the neural network is trained in a self-affirmation manner towards its own predictions, where ground truth labels, unlike for the original EM, are never reassigned by model predictions.  

Let us introduce some notation. Let
\begin{equation} \label{eq:labeleddata}
  (X,G) := \{ (x_j,g_j) \, : \,  j=1,\ldots,N \} \subseteq \mathbb{R}^n \times \mathbb{G}
\end{equation}
denote the collection of input samples, where $X$ denotes the data and $G$ the set of all associated labels from a finite label space $\mathbb{G}$ containing $C$ classes, which in the following are identified with numbers, i.e., $\mathbb{G} := \{1,\ldots,C\}$. We denote the deterministic probability distribution on $\mathbb{G}$ by $\delta_g$, i.e., $\delta_g(g)=1$ and $\delta_g(c)=0$, $g\not=c$. Further, let $X^\prime=\{x'_j\,:\, j=1,\ldots,N'\}$ with $N'\gg N$ denote input samples where no labels are available, and let $f: \mathbb{R}^n \times \mathbb{R}^p \rightarrow \mathbb{Y}$ denote the neural network function where $p$ is the number of learnable parameters and $\mathbb{Y} = \{ y \in [0,1]^C \subseteq \mathbb{R}^C \,:\, \sum_{c=1}^C y_c = 1 \}$ the space of classification distributions. For an input $x \in \mathbb{R}^n$ and weights $w \in \mathbb{R}^p$ we denote the softmax output of the neural network by $\hat{y} = f(x,w)$.

The loss function in our approach is the negative maximum likelihood of the softmax classification rule 
\begin{equation} \label{eq:loss1}
  \mathcal{L}_1(g,\hat{y}) = - \sum_{c=1}^{C} \delta_g(c) \log( \hat{y}_c ) = -\log( \hat{y}_g ) \, .
\end{equation}
Similarly, the normalized classification entropy $\mathcal{H}$ is defined by
\begin{equation} \label{eq:entropy1}
  \mathcal{H}(\hat{y}) = - \frac{1}{\log(C)} \sum_{c=1}^{C} \hat{y}_c \log( \hat{y}_c ) \in [0,1]\, .
\end{equation}
An auxiliary loss for the unlabeled data can be defined as follows. Let pseudo-labels be defined by  
\begin{equation} \label{eq:pseudolabel}
  \psi(\hat{y}) := \mathop{\rm arg\,max}_c \{ \hat{y}_c \, : \, c=1,\ldots,C\}\, ,
\end{equation}
i.e., the index with highest classification probability. Then the auxiliary loss can be expressed using the loss function $\mathcal{L}_1$ from \cref{eq:loss1} via
\begin{equation} \label{eq:auxloss1}
  \mathcal{L}_1(\psi(\hat{y}),\hat{y}) = - \sum_{c=1}^{C} \delta_{\psi(\hat{y})}(c) \log( \hat{y}_c ) \, .
\end{equation}
This term reaches its minimum if $\hat{y} = \delta_{\psi(\hat{y})}$, and indeed we have $\mathcal{L}_1(\psi(\hat{y}),\hat{y}) = \mathcal{H}(\delta_{\psi(\hat{y})}) = 0$. Thus, minimizing \cref{eq:entropy1} is conceptually close to minimizing \cref{eq:auxloss1}.
We can now define a combined loss by
\begin{equation} \label{eq:loss}
  \mathcal{L}(g,\hat{y}) = \begin{cases} \mathcal{L}_1(g,\hat{y})  & \text{ for } x \in X \\
\mu \mathcal{L}_1(\psi(\hat{y}),\hat{y}) & \text{ for } x \in X^\prime \, . \end{cases}
\end{equation}
We do not consider the entropy based equivalent of \cref{eq:loss}, as the loss function $\mathcal{L}$ yields additional freedom in the definition of pseudo-labels $\psi(\hat{y})$ (cf.~\cref{eq:pseudolabel}). The choice of the regularization parameter $\mu$ and other practical implementation details are discussed in \cref{sec:exp_illu,sec:exp}. Having ingredients for the prescription of the EM algorithm with pseudo-labels, we can state a generic version of it in \cref{alg:EM1}, which we specify further in the next paragraph. 

\begin{algorithm2e}[ht]
    \SetAlgorithmStyle
    \caption{Active EM with pseudo-labels\label{alg:EM1}}
    \KwData{$(X,G)$ labeled data from \cref{eq:labeleddata}, $X^\prime$~unlabeled~data}
    \KwResult{weights $w$}
    let $(D,L) := (X,G)$, initialize weights $w$ \;
    train $w$ on $(D,L)$ minimizing $\mathcal{L}$ \; \label{line:inittrain}
    \Repeat{satisfied}{
      \ForAll{$x^\prime \in X^\prime$}{
        infer $y^\prime$ using $f,w$ and $x'$ \label{line:infer} \;
        \If{$y^\prime$ close enough to $\delta_{\psi(y^\prime)}$}{ \label{line:choose}
          $(D,L) \leftarrow (D,L) \cup \{(x^\prime,\psi(y^\prime))\}$ \;
        }
      }
      \For{a chosen number of $x^\prime \in X^\prime$ with $y^\prime$ far away from $\delta_{\psi(y^\prime)}$}{ \label{line:active}
        ask the oracle for the ground truth $g'$ \;
        $(X,G) \leftarrow (X,G) \cup \{(x^\prime,g')\} $ \;
        $x^\prime \leftarrow X^\prime \cup \{ x' \}$ \;
      }
      train $w$ one epoch on $(D,L)$ minimizing $\mathcal{L}$ \;
    }
\end{algorithm2e}

\subsection{Monte-Carlo Dropout Inference} \label{sec:bayesian}

Disregarding the nature of the given data and the prediction task, the practical performance of \cref{alg:EM1} strongly depends on three factors, the initial accuracy achieved in \cref{line:inittrain}, the pseudo-label quality which depends on \cref{line:infer,line:choose}, cf.~e.g.~\cite{DBLP:journals/corr/abs-1710-00209,Lee_pseudo-label:the}, and the acquisition policy in \cref{line:active} for demanding additional ground truth, see~\cite{DBLP:journals/corr/GalIG17}. In this paragraph we focus on the latter two aspects. It has been proposed in \cite{DBLP:journals/corr/abs-1710-00209} to use MC dropout inference for generating pseudo-labels in a semi-supervised setting. In the active learning setting, MC dropout has been used in~\cite{DBLP:journals/corr/GalIG17} to evaluate the uncertainty of a prediction $f(x,w)$ and thus decide which samples to label next by help of an oracle. We combine both approaches as follows.

In~\cref{line:infer} we simply infer using MC dropout with a chosen number $T'$ of forward passes to obtain the average probability outputs
\begin{equation}
  \tilde{y} = \tilde{f}_{T'}(x,w) := \frac{1}{T'}\sum_{t=1}^{T'} f^{(t)} (x,w) \text{ for all } x \in X^\prime \,,
\end{equation}
where $f^{(t)}$ denotes $f$ with dropout. Note, that $f^{(t)}$ is not deterministic when including dropout, i.e., $f^{(t)}$ is not uniquely defined for a chosen $t$.

In order to perform \cref{line:choose} we define a metric that tells us how close $\tilde{y}$ is to $\delta_{\psi(\tilde{y})}$. 
Clearly, many different metrics could be considered. Our metric of choice is the classification entropy $\mathcal{H}(\tilde{y})$ from \cref{eq:entropy1}. For the threshold estimation, we apply MC dropout inference with a chosen number of forward passes $T$ to all available ground truth labeled data $X$ (including data where ground truth is obtained from the oracle during training) and calculate the average classification entropy
\begin{equation} \label{eq:threshold}
  \theta := \frac{1}{|X|} \sum_{x \in X} \mathcal{H}(\tilde{f}_{T}(x,w)) \, .
\end{equation}
We choose a threshold $\theta$ and add in every iteration of \cref{alg:EM1} all samples with entropy below threshold, i.e., $\mathcal{H}(\tilde{f}_{T'}(x',w)) < \theta$. That is,
\begin{equation}
  (D,L) \leftarrow (D,L) \cup \{ (x^\prime,\psi(\tilde{f}_{T'}(x',w))) \} \text{ for } x' \in X' \, .
\end{equation}

For the active learning part in \cref{line:active} we use the entropy of averaged classification results under MC dropout, i.e., $\mathcal{H}(\tilde{f}_{T'}(x^\prime,w))$ for all $x^\prime \in X^\prime$. For a chosen number of samples $x^\prime \in X^\prime$ with highest entropy we ask the oracle for the ground truth $g'$ and add the labeled data $(x',g')$ to $(X,G)$ while removing $x'$ from $X'$. Other approaches for acquiring labels are proposed in~\cite{DBLP:journals/corr/GalIG17}, where it has been shown that the classification entropy under MC dropout is one of the best choices among the considered acquisition functions.

Parameter values for $T$ and $T^\prime$ are stated in the experiments in \cref{sec:exp}.

\section{An Illustrative Example} \label{sec:exp_illu}

\begin{figure*}[htb]
\centerline{ \includegraphics[width=0.24\textwidth]{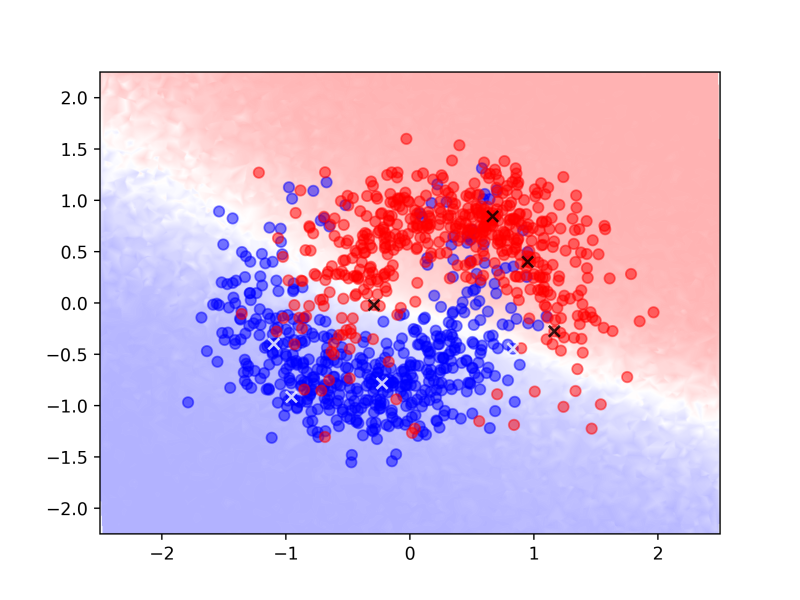} 
 \hfill \includegraphics[width=0.24\textwidth]{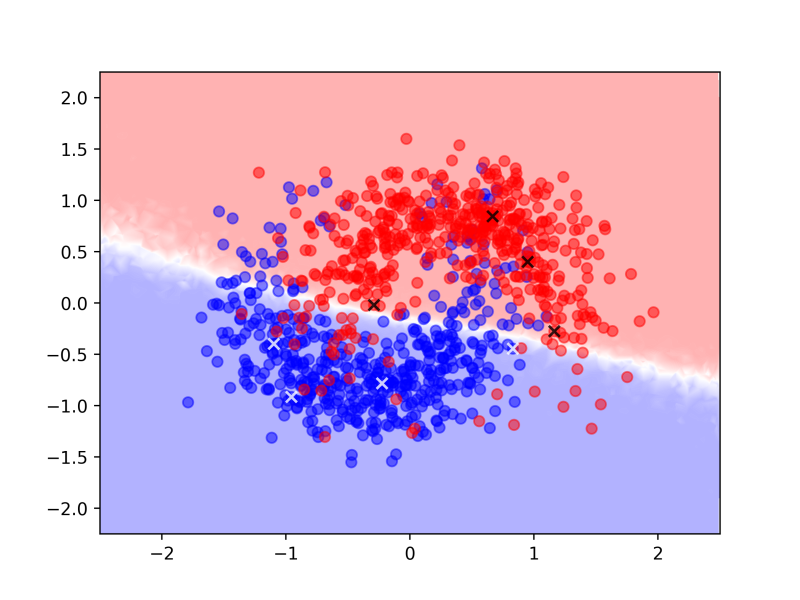}
 \hfill \includegraphics[width=0.24\textwidth]{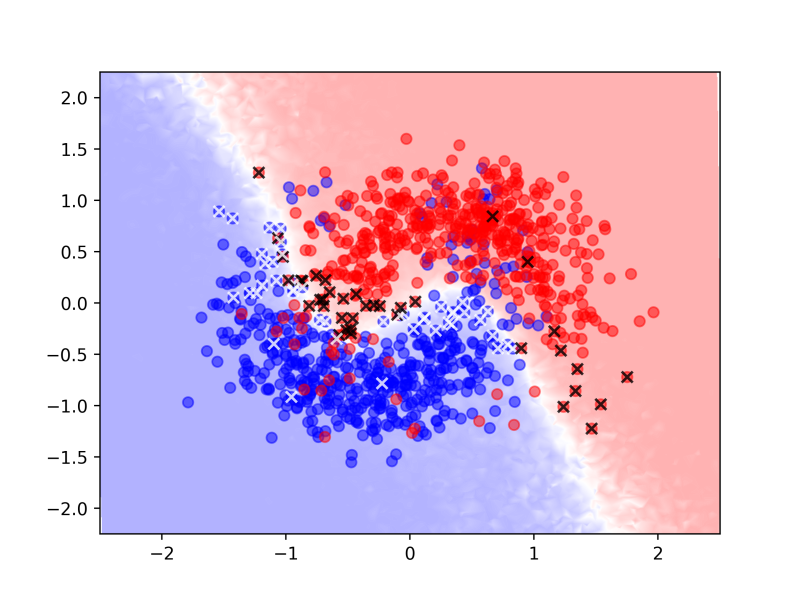} 
 \hfill \includegraphics[width=0.24\textwidth]{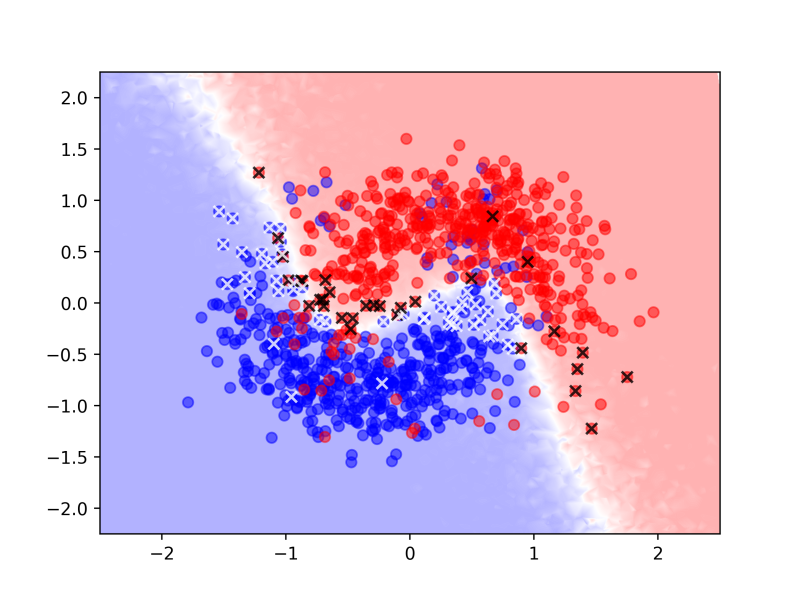}}
\caption{Experiments for \cref{alg:EM1}: \emph{(left)}: training with 8 labels, 4 per class, \emph{(middle left)}: 8 labels + pseudo-labels for the rest, \emph{(middle right)}: active learning with initial 8 labels + 72 over time, \emph{(right)}: active semi-supervised learning with initial 8 labels, pseudo-labels for the rest and 72 labels over time. \label{fig:dist1}}
\end{figure*}

\begin{figure*}[htb]
\centerline{ \includegraphics[width=0.24\textwidth]{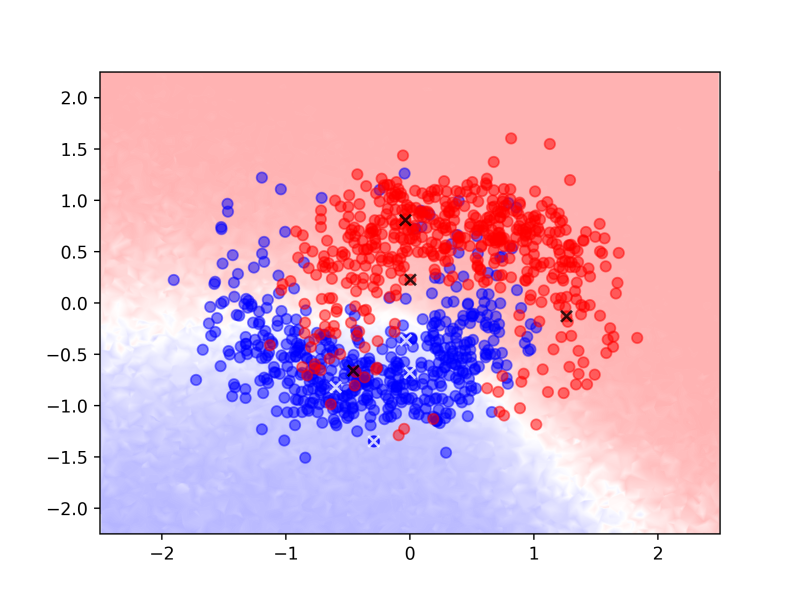} 
 \hfill \includegraphics[width=0.24\textwidth]{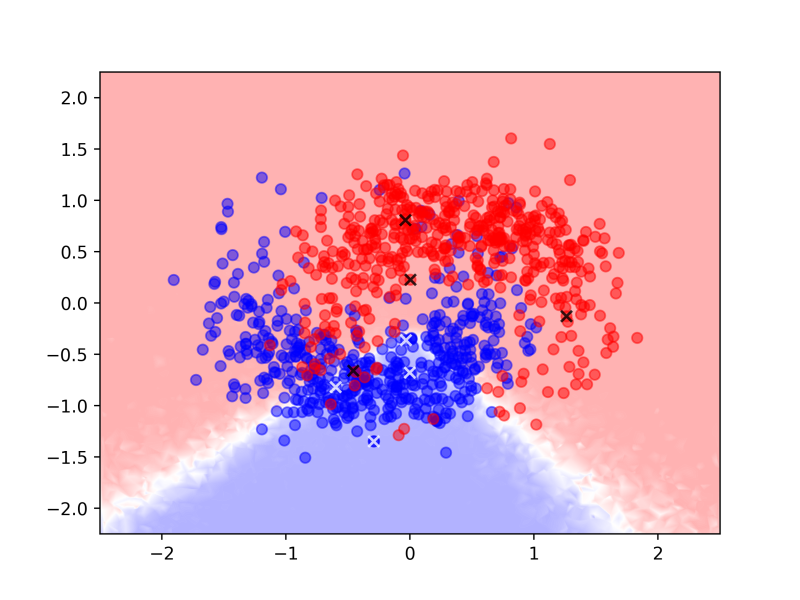}
 \hfill \includegraphics[width=0.24\textwidth]{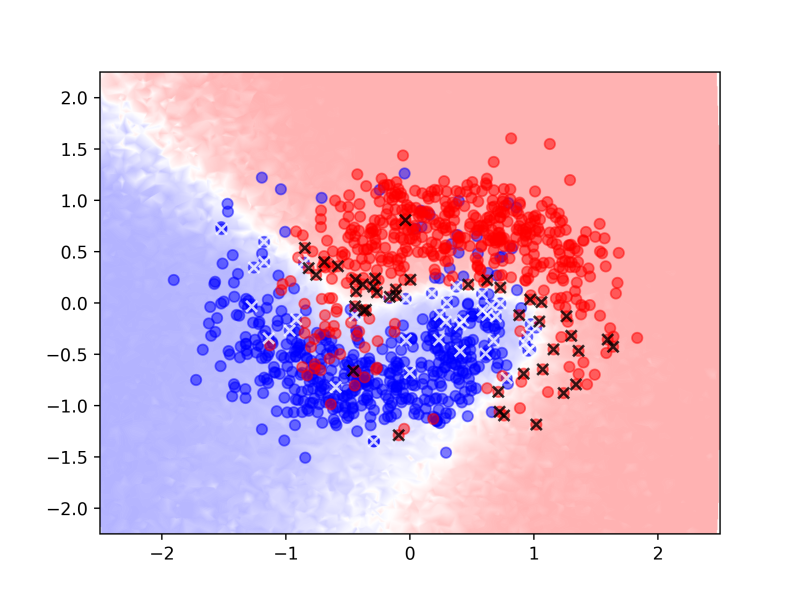} 
 \hfill \includegraphics[width=0.24\textwidth]{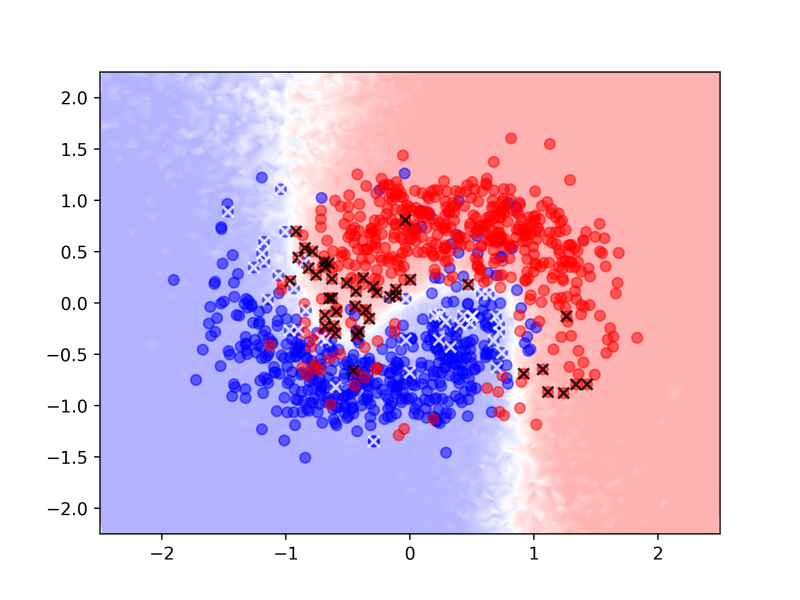}}
\caption{Experiments for \cref{alg:EM1}: same tests as in \cref{fig:dist1}, but a different choice of initial labels.\label{fig:dist2}}
\end{figure*}

In this section we show some experiments with \cref{alg:EM1} for illustration and motivation. Our simple problem consists of $2$ non-convex distributions, from which we draw $500$ samples each. The distributions are generated as follows:
Let $\mathcal{N}(m,\sigma^2)$ be a Gaussian normal distribution with mean $m$ and variance $\sigma^2$. We draw radius and angle from
\begin{equation}
  r \sim \mathcal{N}(1,(1/4)^2) \quad \text{and} \quad \phi \sim \mathcal{N}(1/2,(1/3)^2)  \, .
\end{equation}
Samples from the red distribution are constructed as
\begin{equation}
  \mathit{red} = (1/3,-1/10) + r \cdot ( \cos(\phi), \sin(\phi) )
\end{equation}
and samples from the blue one are constructed as
\begin{equation}
  \mathit{blue} = (-1/3,1/10) + r \cdot ( \cos(\phi), -\sin(\phi) ) \, ,
\end{equation}
together they form a Yin-and-Yang type of picture, see~\cref{fig:dist1,fig:dist2}.

\paragraph{Network Architecture and Parameters}
There are two competing objectives at work when considering a suitable network architecture for deepBASS.
On one hand, we have to employ strong regularization in order to avoid overfitting when learning the small set of initially labeled samples.
On the other hand, the initial model has to re-adjust to new ground truth obtained from the oracle during the EM iteration, which requires flexibility. 

Thus in addition to strong regularization our model needs to be equipped with enough learnable parameters. Hence, for all tests shown in this section we use a fully connected neural network with $2$ input neurons, $3$ hidden layers with $50$ neurons each and $2$ output neurons. After each hidden layer we use the LeakyReLU activation function, i.e.,
\begin{equation} \label{eq:leakyrelu}
  \mathit{LeakyReLU}(x) = \begin{cases} x & \text{ for } x>0 \\ \; 0.1x & \text{ else, } \end{cases}
\end{equation}
followed by dropout~\cite{JMLR:v15:srivastava14a} with $33\%$ dropout rate.
All models are trained and evaluated using Keras~\cite{chollet2015keras} with the Tensorflow backend~\cite{tensorflow2015-whitepaper}.
All layers are $L^2$ regularized with regularization parameter $\lambda=10^{-3}$. To fit the weights we use Adam~\cite{DBLP:journals/corr/KingmaB14} with default parameters. The mini-batch size is $256$.

For MC dropout inference on unlabeled and pseudo-labeled data $X^\prime$ we use $T'=10$ forward passes with dropout, and on the data $X$ with known ground truth we perform $T=100$ forward passes. In every iteration of \cref{alg:EM1} unlabeled data from $X^\prime$ that is not in $D$, yet, is added if its classification entropy is below the threshold $\theta$, cf.~\cref{eq:threshold}.

\paragraph{Intentional Overfitting of Ground Truth}
The choice of the regularization parameter $\mu$ in the loss function in \cref{eq:loss} plays an important role. Chosen too small, the use of unlabeled data will barely have any effect, chosen too large, the classification accuracy (rate of correctly predicted classes) on the ground truth labeled data will decrease while the iteration in \cref{alg:EM1} proceeds. Thus we choose to rather overfit the scarce ground truth data in order to allow \cref{alg:EM1} to find a clustering of the unlabeled data $X'$ consistent with the labeled data $X$.

In our Keras implementation, we implemented this balance by ``upsampling'' $X$ in $D$, i.e., a sample from $X$, with known ground truth, is contained $20$ times in $D$ while a sample from $X^\prime$ will be contained (at most) once in $D$. E.g.~for $8$ labeled samples and $992$ unlabeled ones, we can expect about $14\%$ of the data in a mini-batch to be labeled with ground truth. In our tests we observe that this leads to overfitting the ground truth, however it prevents the neural network from forgetting the crucial information.

\begin{table}[b]
\resizebox{0.48\textwidth}{!}{%
\begin{tabular}{*{3}{l}} %
Method                  & panel in \cref{fig:dist1,fig:dist2} & val.\ acc.\   \\ \hline \hline
initial model           & (left)                              & $83.27\%$     \\
semi-supervised         & (middle left)                       & $84.33\%$     \\
active                  & (middle right)                      & $90.19\%$     \\
active semi-supervised  & (right)                             & $90.33\%$     \\
80 labels               &                                     & $88.65\%$     \\
1,\!000 labels          &                                     & $91.37\%$     \\ \hline \hline
\end{tabular}%
}
\caption{Average classification accuracies for the tests performed in \cref{fig:dist1,fig:dist2}.\label{tab:motres}}
\end{table}

\paragraph{Experiments.}
In all four panels in \cref{fig:dist1,fig:dist2} we start with 4 labels per class, the depicted data points represent all data available for training, i.e., $X \cup X^{\prime}$, the data points in $X$, where ground truth is available, are crossed out. The background color gradients depict where the neural network predicts the red or the blue class, respectively. The classification boundary is white and represents the region where the classification uncertainty is high. \Cref{fig:dist1,fig:dist2} only differ in the choice of the $8$ labels. While the choice in \cref{fig:dist1} is easy to handle, the choice in \cref{fig:dist2} represents a rather ill-posed case.

For the left panel in \cref{fig:dist1,fig:dist2} we train only on the available ground truth until the classification accuracy on the training set stagnates. For each figure, all other panels share the left panel as initial model.

In the middle left panel we continue with semi-supervised training, not adding any further ground truth labels. The resulting model in both cases is more sure about its decision, this is indicated by background colors that are more saturated. However in \cref{fig:dist2} the classification boundary got worse compared to the left panel. Both figures also show that it can happen that semi-supervised learning does not perform well, especially when a low density distribution at the class boundaries is not present.

In the center right panel we use active learning. Every second iteration, we demand two ground truth labels. We keep iterating until $80$ samples are labeled, i.e., $72$ iterations. In the right hand panel we use both, active and semi-supervised learning. While in \cref{fig:dist1} active learning performs just as well as active semi-supervised learning, the latter clearly is superior in \cref{fig:dist2}. The interpretation of these results is that the semi-supervised learning component has a regularizing effect on active learning.

Summarizing these tests, we state classification accuracies averaged over $10$ runs for all four tests
in \cref{tab:motres} and complement these with results for purely supervised learning using $80$ and $1,\!000$ labels, respectively. The corresponding models are trained until validation accuracy stagnates.

\section{Experiments with MNIST} \label{sec:exp}

For our experiments, we use the MNIST dataset~\cite{lecun-98} of handwritten digits, given as tiny $28 \times 28$ gray scale images with the pre-defined data split of $60,\!000$ training and validation $10,\!000$ images. Again, all models are trained and evaluated using Keras with Tensorflow backend. For the CNN architecture we use a generic building block containing the following components:

\begin{itemize}
\setlength\itemsep{-0.25em}
  \item convolutional layer with $16$ filters of size $3 \times 3$,
  \item LeakyReLU activation function, \cref{eq:leakyrelu},
  \item dropout with $33\%$ dropout rate.
\end{itemize}

We stack four of these building blocks, after the second and the fourth layer we apply $2\times2$ max pooling. This results in a $7 \times 7 \times 16$ tensor, followed by a dense layer with $10$ outputs and a final softmax activation. The resulting network is equipped with $14,\!970$ learnable parameters. All convolutional layers are trained with $L^2$ regularization and a regularization parameter $\lambda=10^{-3}$. We again use Adam with default parameters for training.




\paragraph{Parameters.}

Throughout our experiments we use the following parameters. For MC dropout inference on unlabeled and pseudo-labeled data $X^\prime$ we use $T'=10$ forward passes of dropout, and on data $X$ with known ground truth we perform $T=100$ forward passes. In each test we perform $200$ iterations of \cref{alg:EM1} and we perform each test $10$ times while re-sampling the initial $100$ samples. The presented results are averages of these $10$ runs, the ground truth up-sampling factor is $20$.

The initial neural network is trained on a balanced data set containing the same number of samples for each class. By default we start with $100$ labeled samples, i.e., $10$ per class. By presenting the ground truth labeled data $2,\!000$ times we obtain a training accuracy of roughly a $99$--$100\%$. During the $200$ iterations in \cref{alg:EM1} we track the performance by monitoring validation accuracy. 
When we perform active learning, we acquire $10$ labels once every $10$ iterations. Note, that the added labels are not necessarily class-balanced.

\paragraph{Experiments with Entropy Thresholding and Label Acquisition Policy.}

\begin{figure*}[htb]
\centerline{\includegraphics[width=1.0\textwidth]{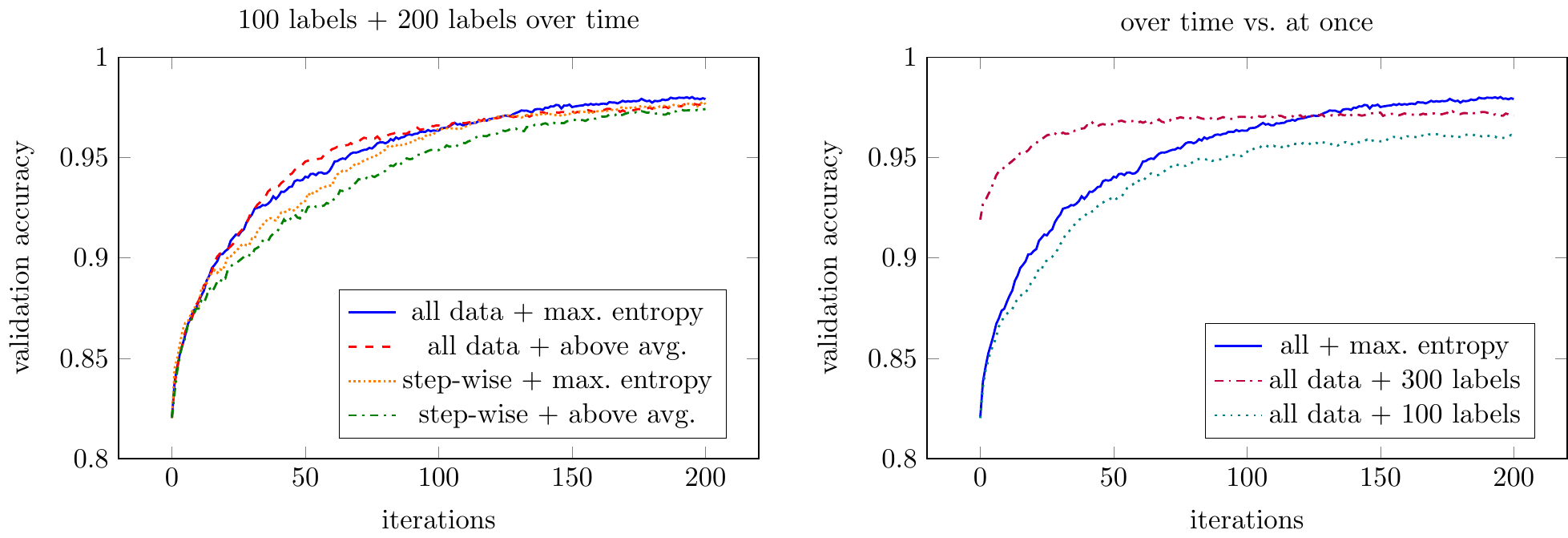}}
\caption{Experiments for \cref{alg:EM1} with two different thresholds for adding data, each with two different label acquisition policies.\label{fig:valacc2}}
\end{figure*}

\Cref{fig:valacc2} shows the behavior of \cref{alg:EM1} over the course of $200$ iterations, averaged over $10$ runs. In the left panel we study the influence of the threshold $\theta$, i.e., we compare the case where all unlabeled data is used in training right from the start ($\theta=1$, short hand: \emph{all data}) with the strategy where pseudo-labeled data is added step-wise according to the threshold $\theta$ from \cref{eq:threshold} (short hand: \emph{step-wise}).
This comparison is made while using two different label acquisition policies. On one hand we ask for labels of unlabeled samples $x^\prime \in X^\prime$ with maximum entropy as explained in \cref{sec:bayesian} (short hand: \emph{max.~entropy}), on the other hand we try a slightly more careful policy where we only demand labels for samples $x^\prime \in X^\prime$ randomly drawn from all data in $X^\prime$ with entropy above average (short hand: \emph{above avg.}), i.e.,
\begin{equation}
  \mathcal{H}(\tilde{f}_{T'}(x^\prime,w)) >  \frac{1}{|X|+|X^\prime|} \sum_{x \in X \cup X^\prime} \mathcal{H}(\tilde{f}_{T'}(x,w)) \, .
\end{equation}
The latter strategy is motivated by the fact that it might happen that exclusively acquiring data with high classification entropy could result in overfitting (wiggly decision boundaries) of data from non-separable distributions, consequently slowing down the convergence of \cref{alg:EM1}.

In our tests with $100$ initial ground truth labels, all four approaches share the same initial models in each run, the average validation accuracy is $82.05\%$ after training with $100$ labeled samples evenly distributed over all classes. The left panel of \cref{fig:valacc2} shows that the \emph{all data} + \emph{max.\ entropy} approach is slightly superior to the same acquisition policy where unlabeled data is added \emph{step-wise}. We believe that the reason for this is in part the well-behaved nature of the MNIST dataset. The \emph{above avg.} acquisition policy is slightly inferior, but works better when all available unlabeled data is used from the beginning.
Summarizing,  \emph{all data} combined with \emph{max.\ entropy} acquisition policy reaches $97.92\%$ accuracy on average. This result is the average of all $10$ runs stopping after $200$ iterations and measuring the accuracy of the model after the last iteration. The best result in a single run is $98.33\%$. In contrast to this, the most careful approach, \emph{step-wise} + \emph{above avg.}, ends up with $97.43\%$ which is still good.

In order to understand how much we benefit from combining active learning and semi-supervised learning, we compare the best approach from the left panel of \cref{fig:valacc2} with \cref{alg:EM1}, but without the active learning component. That is, we use $100$ and $300$ labels from the beginning and perform semi-supervised learning without adding any further ground truth labels. The results are depicted in the right hand panel of \cref{fig:valacc2} and they show that active learning is indeed beneficial when using \cref{alg:EM1} for semi-supervised learning. The pure semi-supervised approach with $300$ labels ends up with an average accuracy of $97.10\%$ which is $0.82\%$ less than for active semi-supervised learning. Note, that we achieve $96.08\%$ with semi-supervised learning and $100$ labels. All results from this section are summarized in \cref{tab:plotres}, complemented with result for supervised and semi-supervised learning with different numbers of labels. Compared to the \emph{all data} + \emph{max.\ entropy} approach, pure supervised learning with a random sample of labeled data requires about $10$ times as many labels. The samples standard deviation reveals that our approach is robust under data re-sampling.

\begin{table}[tb]
\resizebox{0.48\textwidth}{!}{%
\begin{tabular}{*{4}{l}} %
Threshold     & policy          & ground truth        & val.~acc.(stddev.) $\%$ \\ \hline \hline
--            & --              & $100$               & $82.03(\pm1.95)$     \\
--            & --              & $300$               & $91.91(\pm0.75)$     \\
--            & --              & $600$               & $94.92(\pm0.32)$     \\
--            & --              & $1,\!000$           & $96.24(\pm0.39)$     \\
--            & --              & $3,\!000$           & $97.89(\pm0.25)$     \\ \hline
step-wise     & max.~entropy    & $300$               & $97.67(\pm0.34)$     \\
step-wise     & above average   & $300$               & $97.43(\pm0.19)$     \\
all data      & max.~entropy    & $300$               & $97.92(\pm0.19)$     \\
all data      & above average   & $300$               & $97.65(\pm0.26)$     \\ \hline
all data      & --              & $100$               & $96.08(\pm1.49)$     \\
all data      & --              & $300$               & $97.10(\pm0.40)$     \\
all data      & --              & $600$               & $97.39(\pm0.35)$     \\
all data      & --              & $1,\!000$           & $97.64(\pm0.23)$     \\
all data      & --              & $3,\!000$           & $98.14(\pm0.26)$     \\ \hline
--            & --              & $60,\!000$          & $99.09(\pm0.07)$     \\ \hline \hline
\end{tabular}%
}
\caption{Summary of average classification accuracies for the tests performed in \cref{fig:valacc2}. For all tests with label acquisition policy, $100$ ground truth labels are used initially, $200$ are added over time.\label{tab:plotres}}
\end{table}

\paragraph{Data Augmentation.}

Except for this paragraph, all tests in this work are performed without data augmentation. However in a practical setting it might make sense to use data augmentation as well. For the MNIST dataset, when using data augmentation on the ground truth labeled data with slight rotations of less than $10$ degrees and slight image scaling of up to $5\%$ in height and width, we observe that $50$ labels are enough to achieve competitive initial validation accuracies of around $85\%$.

\paragraph{Comparison with Other Methods.}

In this section, we provide an overview of methods for semi-supervised deep learning and active deep learning where tests with the MNIST dataset have been performed. Most of the referred works provide numbers for $1,\!000$ labels, results for $300$ labels are scarce.
We compare these results with the \emph{all data} + \emph{max.\ entropy} Deep Bayesian Active Semi-Supervised learning approach.
For comparison we run our method $10$ times until $1,\!000$ samples are ground truth labeled and average over all validation accuracies, achieving $98.94\%$ validation accuracy. A comprehensive comparison is stated in \cref{tab:comp}. Clearly our approach, using $1,\! 000$ labeled samples is competitive at the upper end of the spectrum of reported results. Though one of the main advantages of it, as reported in the previous sections, is its ability to yield high accuracies even with as few as only $300$ labeled samples. 
Note, that the full ladder net model from~\cite{DBLP:RasmusVHBR15} is a very sophisticated model incorporating denoising auto-encoder structures which might lack scalability and portability.

The semi-supervised part of our method with only $100$ labels reaches a validation accuracy of $96.08\%$, a similar approach without MC dropout, see~\cite{Lee_pseudo-label:the}, only achieved $89.51\%$. We observed similar results in our tests without MC dropout inference which reveals its impact.

\begin{table}[ht]
\resizebox{0.48\textwidth}{!}{%
\begin{tabular}{*{3}{l}} %
Method                                                                 & test error  \\ \hline \hline
Semi-Supervised:                                                       &             \\ \hline
\cite{Weston2012}: Semi-Supervised Embedding                           & $5.73\%$    \\ 
\cite{Weston2012}: Transductive SVM                                    & $5.38\%$    \\
\cite{10.1007/978-3-662-44851-9_36}: AtlasRBF                          & $3.68\%$    \\
\cite{NIPS2011_4409}: Manifold Tangent Classifier                      & $3.64\%$    \\
\cite{Lee_pseudo-label:the}: Pseudo-label                              & $3.46\%$    \\
\cite{DBLP:journals/corr/abs-1710-00209}: Self training + Dyn.~conf.\  & $3.42\%$    \\
\cite{DBLP:journals/corr/KingmaRMW14}: Deep Generative Models          & $2.40\%$    \\ 
\cite{DBLP:RasmusVHBR15}: Ladder, $\Gamma$-model                       & $1.53\%$    \\
\cite{DBLP:conf/icml/HuMTMS17}: Virtual Adversarial                    & $1.32\%$    \\
\cite{DBLP:RasmusVHBR15}: Ladder, full                                 & $0.84\%$    \\ \hline
Active:                                                                &             \\ \hline
\cite{DBLP:journals/corr/GalIG17}: Bald                                & $1.80\%$    \\
\cite{DBLP:journals/corr/GalIG17}: Max Entropy                         & $1.74\%$    \\
\cite{DBLP:journals/corr/GalIG17}: Var Ratios                          & $1.64\%$    \\ \hline
Active + Semi-Supervised:                                              &             \\ \hline
DeepBASS (\emph{all data} + \emph{max.~entropy}):                      & $1.06\%$    \\ \hline \hline
\end{tabular}%
}
\caption{Comparison with other approaches for a $1,\!000$ labels. We term our method  \label{tab:comp}}
\end{table}

\section{Conclusion \& Outlook}

We have introduced a general active semi-supervised deep learning method with a wide field of possible applications that shows great performance in first results for the MNIST dataset. While we use only simple classification entropy based uncertainty quantification, the presence of approximate Bayesian inference as well as the combination of semi-supervised learning and active learning constitute to the strength of our method as it outperforms state-of-the-art general approaches which do not use advanced network architectures.

If validation data is available, our approach can be further tuned with respect to thresholding and acquisition policy. This fact implies, that additional meta-learning extensions could be developed. A minor concern might be, that data which is added in the active part of the approach is prone to overfitting. A clean restart with the final data splitting and further tuning could additionally improve the performance of our method.

We plan to produce results for this approach in different applications and provide our source code on GitHub, cf.\ \texttt{\href{https://github.com/mrottmann/DeepBASS}{https://github.com/mrottmann/DeepBASS}}.

\paragraph{Acknowledgements.}

We would like to thank Fabian H\"uger and Peter Schlicht from Volkswagen Group Research for discussion and remarks on this work.


{
\printbibliography
}


\end{document}